\documentclass[a4paper]{article}
\usepackage{INTERSPEECH2020}
\usepackage{multirow}
\usepackage{graphicx}
\usepackage{lscape}
\usepackage{float}
\title{A Full Text-Dependent End to End Mispronunciation Detection and Diagnosis with Easy Data Augmentation Techniques}
\name{Kaiqi Fu$^1$, Jones Lin$^1$,Dengfeng Ke$^1$,Yanlu Xie$^1$,Jinsong Zhang$^1$,Binghuai Lin$^2$}
\address{
  $^1$Beijing Language and Culture University\\
  $^2$Smart Platform Product Department,Tencent Technology Co., Ltd, China }
\email{ \{kaiq.fu, jonneslin\}@gmail.com, \{dengfeng.ke, xieyanlu, jinsong.zhang\}@blcu.edu.cn, binghuailin@tencent.com}

\begin{document}

\maketitle
\begin{abstract}
Recently, end-to-end mispronunciation detection and diagnosis (MD\&D) systems has become a popular alternative to greatly simplify the model-building process of conventional hybrid DNN-HMM systems by representing complicated modules with a single deep network architecture. In this paper, in order to utilize the prior text in the end-to-end structure, we present a novel text-dependent model which is difference with sed-mdd, the model achieves a fully end-to-end system by aligning the audio with the phoneme sequences of the prior text inside the model through the attention mechanism. Moreover, the prior text as input will be a problem of imbalance between positive and negative samples in the phoneme sequence. To alleviate this problem, we propose three simple data augmentation methods, which effectively improve the ability of model to capture mispronounced phonemes. We conduct experiments on L2-ARCTIC, and our best performance improved from 49.29$\%$ to 56.08$\%$ in F-measure metric compared to the CNN-RNN-CTC model. 

\end{abstract}


\section{Introduction}

With the development of speech technology and the promotion of online learning, computer-aided pronunciation training (CAPT) has been more fully applied to language teaching \cite{neri2008effectiveness,Xie2020}.  As an important part of CAPT technology, mispronunciation detection and diagnosis (MD\&D) can be used to detect the mispronunciation in a L2 learner‘s speech, and further diagnose and give learners effective feedback.

A great deal of research on mispronunciation detection has been carried out [3-9], these methods can be grouped into two categories. The first are pronunciation scoring based on confidence measures originally proposed for automatic speech recognition (ASR). A namely goodness of pronunciation (GOP) scores \cite{Witt} is computed by log-posterior probability based on force alignment, the method and its prominent variants \cite{hu,CaGOP} currently mostly popular and achieve a promising performance on mispronunciation detection. But, this kind of method is not only unable to deal with the insertion errors in pronunciation, but also fail to give more detailed diagnosis to learners. 

The second category aims to assess the type of the mispronunciation and provide informative feedback on specific errors. A well-know method of this category is extend recognition network (ERN) \cite{Harrison2009ImplementationOA}, which incorporate expected pronunciation error patterns into the lexicon to constrain the recognition paths to the canonical pronunciation and the likely phonetic mispronunciations. These ERNs \cite{Harrison2009ImplementationOA,ERN,qianern} compiled by handcrafted or data-driven rules \cite{meng2007deriving} have the advantage that the errors and the error types are detected together, and thus can be used for the system to provide diagnostic feedback. But it is difficult to build ERNs that incorporate as many as possible mispronunciation paths so that the recall performance improvement is limited \cite{leung2019cnn}.

Recently, the end-to-end structure shows a good talent for ASR tasks \cite{graves2014towards,amodei2016deep}, and gets promising performance in MDD \cite{leung2019cnn,feng2020sed,yan2020end}. A CNN-RNN-CTC free phone recognize model was proposed \cite{leung2019cnn} firstly, which showed an outperformed result comparied with previously approaches \cite{li2016mispronunciation}. The hybrid CTC-Attention architecture \cite{yan2020end} with expanding the original L2 phone set was used to detect categorial and non-categorial errors. Those CTC-based methods does not require forced alignment, and intergrate the whole training pipeline. However, in the case of the reading text already known, the above research does not use the existing prior text information. To utilize the prior linguistic information, previous study \cite{feng2020sed} proposed a text-dependent end to end model that uses attention mechanism to combine the high-level hidden features of acoustics  and the sentence-to-character encoder feature. 

The model proposed in this paper is similar to sed-mdd \cite{feng2020sed} to utilize the prior text information, but there are still some differences. Feng et.al convert the prior text into a sequence of characters for sentence-embedding, The task of MD\&D aims to detect errors in phoneme-level, so it makes sense to convert text into phoneme sequences and fed them into the sentence encoder. In addition, we also noticed that in the sed-mdd, manually labeled phoneme boundary was used to calculate the frame-level cross-entropy loss function. In this paper, connectionist temporal classification (CTC) \cite{graves2006connectionist} loss function was used without any labeled time information.

In this work, we propose an end-to-end mispronunciation detection framework based on the prior text attention mechanism, and explore the effects of sentence-to-phoneme. Furthermore, there are many more correctly pronounced phonemes than incorrectly pronounced in the training set, there will be a problem of imbalance between positive and negative samples in the phoneme sequence we send to the attention model, which leads to the tendency to output the input phoneme sequences. For this we propose three easy data augment techniques to explore the influence of modifying the input phoneme sequences. Finally, in our experiment, all datasets, metrics and baseline systems are open source$\footnote{https://github.com/cageyoko/CTC-Attention-Mispronunciation}$.

The rest of this paper is organized as follows. The following section describes our system in detail and introduces several methods of data augmentation. In section 3, experimental results will demonstrate. Section 4 concludes with a discussion of potential future work.

\begin{figure}[t]
  \centering
  
  \includegraphics[width=\linewidth]{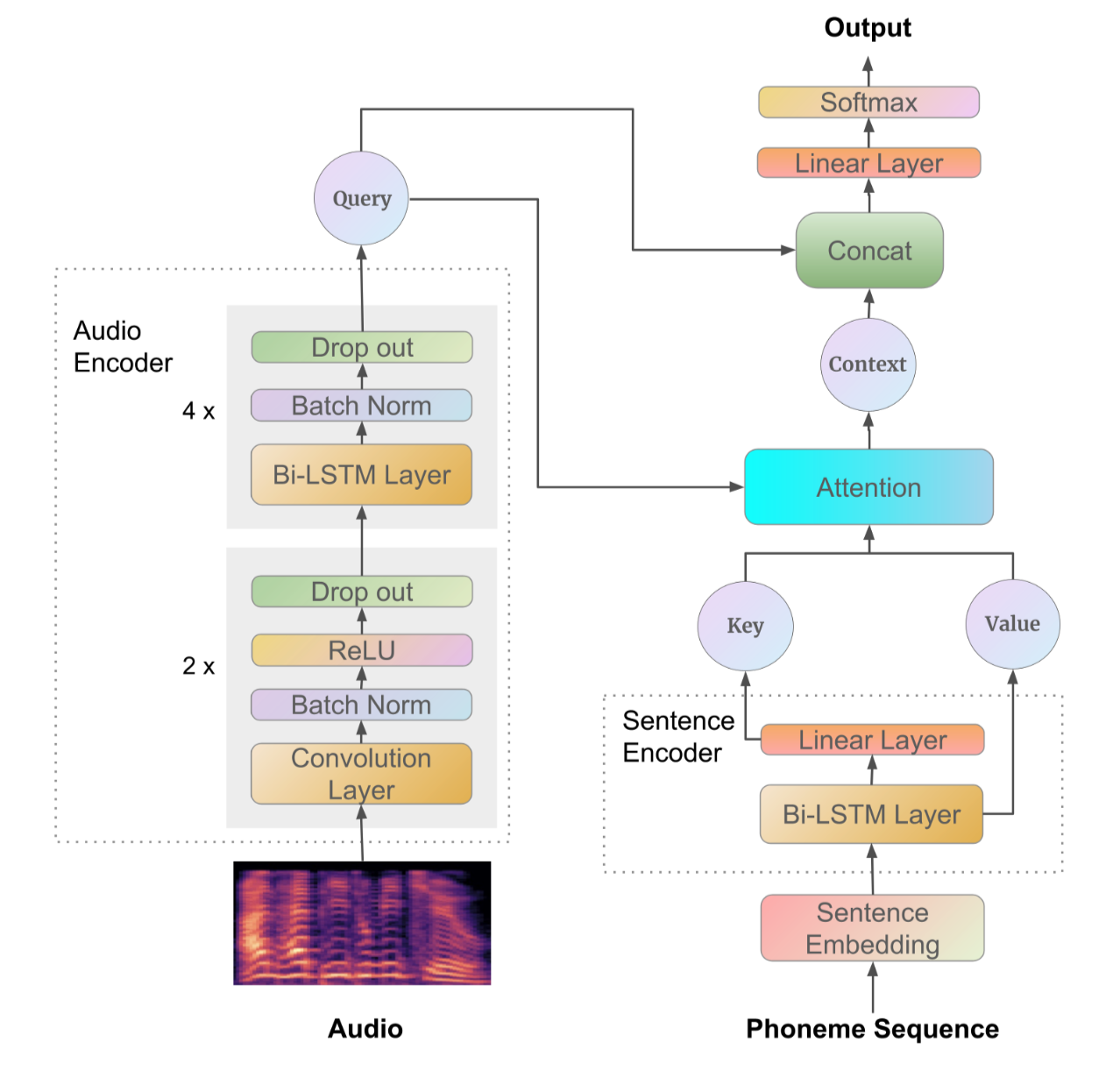}
  \caption{The proposed model structure}
  \label{fig:model}
\end{figure}

\section{Our Methods}
\subsection{System overview}

The input of the model is the phoneme sequence converted from prior text and the fbank acoustic feature, and the output of the model is the phoneme sequence corresponding to the audio. Specifically, it is assumed that the audio input is $x = [x_1, ... ,x_t, ... ,x_T]$, where $x_t$ represents the feature vector of speech at frame $t$ and $T$ represents the total number of frames of a speech. The $N$-length phoneme sequence input is $s = [s_1, ... , s_n , ...,s_N]$, $s_n$ is a phoneme at position $n$ in the prior phoneme sequence. Our model contains three modules: the sentence encoder, the audio encoder and decoder with the attention mechanism. The two encoders extract high-level feature representations $h$ from the input speech feature $x$ and sentence-to-phoneme sequences $s$: 
\begin{equation}
   h^Q = \mathrm{Audio\_encoder(}x) \label{eq:audio_encoder}
\end{equation}
\begin{equation}
   h^K, h^V = \mathrm{Sentence\_encoder}(s) \label{eq:sentence_encoder}
\end{equation}
where the $h^Q$ and $h^K$ are the query and key. In our work, wo adopt CNN-RNN struct as the audio encoder that has been experimented in the previous research \cite{leung2019cnn} and showed a good results, Bi-LSTM as the sentence encoder that has strong sequential modeling ability. Then the attention mechanism takes the $h^Q$,$h^K$ and $h^V$ as input and forms a fixed-length context vector:
\begin{equation}
    c = \mathrm{Attention}(h^Q, h^K, h^V) \label{eq:attention}
\end{equation}

The workflow of the entire model is shown in Figure~\ref{fig:model}. we detail the encoders and the attention mechanism in the following subsections.
  
\subsubsection{Sentence Encoder}
The input of the encoder is phoneme sequences of the prior text,  after feeding the phoneme sequence $s$ into the model, each phoneme is embedded into a vector and fed into a Bi-LSTM with hidden size=384 and dropout=0.2 to obtain the output sequence $ { [h{^V_1}, ... ,h{^V_{n}}, ... ,h{^V_{N}}] }$, denoted as value, where $h{^V_{n}}$ is 2*hidden size. This value is fed into linear layer with 2*hidden size as input and 2*hidden size as output to obtain the output sequence $ [h{^K_1}, ... ,h{^K_{n}}, ... ,h{^K_{N}}]$, denoted as key, where $h{^K_{n}}$ is equal to 2*hidden size. the key and value will be fed into the decoder module in the next step.
\begin{equation}
  e_n \mathrm{ = Sentence\_Embeding(}s_n)  \label{eq:embed}
\end{equation}
\begin{equation}
   h{^V_n} = f_1(e_n) \label{eq:hv}
\end{equation}
\begin{equation}
   h{^K_n} = f_2(h{^V_n}) \label{eq:hk}
\end{equation}
where $f_1(.)$ represents function of Bi-LSTM cell and $f_2(.)$ represents fully connected layer.

\subsubsection{Audio Encoder}
The input of the encoder is a 243-dimensional feature. It is worth noting that the stacking of the left and right frames and the current frame is used here, and the original audio feature is an 80-dim fbank and 1-dim energy. After the audio features $x_t$ are fed into the model, they first go through the CNN-RNN module to get the output sequence $ [h{^Q_1}, ..., h{^Q_{t'}}, ...,h{^Q_{T'}}] $, denoted as query. $T'$ denotes the length of the original $T$-frame speech after the CNN downsampling operation, where $T' \leqslant T$.
\begin{equation}
  {   h{^Q_{t'}} = f_3(x_t) } \label{eq:hQ} 
\end{equation}
where $f_3(.)$ represents function of CNN-RNN layers. The CNN-RNN module consists of two CNN layers and four Bi-LSTM layers. Batch normalization is employed following each layer to help the model converge better.

\subsubsection{Decoder with attention}

Recall that the attention mechanism forms a context vector c that 
contains the information of key $h^K$ . Notably, the attention mechanism can use both past and future frames of a time sequence. Thus, a normalized attention weight is learned using $h{^Q_{t'}} $ and $  h{^K_n}$:
\begin{equation}
   \mathrm { \alpha{_{t',n}} = \frac { exp(score(h{^K_n},h{^Q_{t'}}))} {\sum\nolimits_{n=1}^N exp(score(h{^K_n},h{^Q_{t'}})) } } \label{eq:alpha}
\end{equation}
where $\mathrm { score(h{^K_n},h{^Q_{t'}}) = {h{^Q_{t'}}}h{^K_n}^T }$. $\alpha$ is the attention weight, which makes a monotonic alignment between the audio and the prior text input. Finally, we compute the context vector as the weighted average of $h^V_n$: 
\begin{equation}
   \mathrm {  c_{t'} = { \sum_{t'=1}^{T'} \alpha{_{{t'},n}}h{^V_n} } } \label{eq:cvector}
\end{equation}

Considering that the text information used may be too strong in the model, here we use the previous value $h^Q$ as acoustic residual. Given $c_{t'}$ and $h^Q_{t'}$, the final framewise probability $y_{t'}$ is:
\begin{equation}
   y_{t'} = \mathrm{softmax}( W[c_{t'};h^Q_{t'}] + b )
\end{equation}
where the [·;·] denotes the concatenation of two vectors. Beamsearch is used here to generate recognized phoneme sequence with $y_{t'}$.
\subsection{Data Augmentation}
One difficult of the model our proposed is the imbalance between positive and negative samples in training data. Here, we define a positive sample as a mispronunciation phone, and negative samples as one that is the canonical. Without data augmentation, the model would tend to output the prior phonemes and ignore the learn‘s mispronunciation, i.e., higher accuracy and lower recall, because the prior phoneme sequences were fed into the model, the most of which is negative samples, i.e., the number of negative samples is high and the number of positive samples is less, which leads to the model preferring to output canonical phonemes.  A key goal of mispronunciation detection is to detect the wrong phonemes, thus to make our model can learn more features of positive samples, we increase the number of positive samples by the following three easy random replacement data augmentation operations:

\noindent\textbf{1. Phoneme set based(PS):} Randomly choose phonemes from phoneme sequence of the prior text, replace each of these phonemes with the phoneme set at random, e.g. $ /eh/\to/hh/$. It is worth noting that "blank" symbol may be used instead of the phoneme, which is equivalent to an INSERT type error. It may also happen that the "blank" is replaced by a phoneme in the reading text, which is equivalent to a DELETE type error. 

\noindent\textbf{2. The vowels and consonants set based(VC):} The vowels are more likely to be mispronounced with vowels and consonants with consonants based on L2-ARCTIC statistics \cite{l2arctic}  such as the frequent substitution: $ /z/\to/s/$. So randomly choose phones from phoneme sequences of the prior text, if the phoneme belong to vowel/consonant, replace it with a phoneme in vowel/consonant set at random.

\noindent\textbf{3. The confusing pairs based(CP):} Firstly, we count the confused pairs of learner pronunciation on the phonemes in the L2-Arctic part of the training set, then, we randomly choose phones from phoneme sequences of the reading text, if the phoneme belong to confused pairs, replace it with with its confusing phoneme at random.

\section{EXPERIMENTS}
\subsection{Speech Corpus}
We conducte TIMIT \cite{timit} and L2-ARCTIC \cite{l2arctic}{} datasets to evaluate the performance of the our model, both corpora are publicly available. The L2-ARCTIC is a non-native English corpus built for CAPT and other tasks, it contains 24 non-native speakers (12 males and 12 females), whose L1 languages include Hindi, Korean, Mandarin, Spanish, Arabic and Vietnamese.

To unify the phone set of these two datasets, we map the 61-phone set in TIMIT and the 48-phone set in L2-Arctic to the same 39-phone set. All TIMIT data are used as training data. for L2-ARCTIC that has been annotated by experts, we select 6 speakers ("NJS", "TLV", "TNI", "TXHC", "YKWK", "ZHAA") to build the test set, and the 6 speakers ("MBMPS", "THV", " SVBI", "NCC", "YDCK", "YBAA") are used as development data to save our best model, remainding 12 speakers as training data.  The detail of data split is shown in Table~\ref{tab:data split}.


In our experiments, the input phoneme sequences were provided by these two corpora. Certainly, we also can produce it by the Montreal forced-aligner \cite{forcealign} as \cite{l2arctic}. Our all models are trained using same parameters, such as learning rate, batch size, etc. 

\begin{table}[H]
\caption{Details of the experimental speech corpora.}
    \centering
    \renewcommand\arraystretch{1.5}
\begin{tabular}{|l|c|c|c|c|} 
\hline
           & TIMIT & \multicolumn{3}{l|}{~ ~ L2-ARCTIC~ ~}  \\ 
\cline{2-5  }
   & Train & Train & Dev  & Test                    \\ 
\hline
Speakers   & 630 &  12 &  6 & 6                    \\ 
\hline
Utterances & 6300  & 1800  & 897  & 900                     \\ 
\hline
Hours      & 4.5   & 1.84  & 0.94 & 0.88                    \\
\hline
\end{tabular}
    \label{tab:data split}
\end{table}

\subsection{Performance for phone recognition}
In this section, phone recognition evaluation metric employed is the phone error rate(PER), which is computed by aligning the annotated phone sequence and recognized phone sequence by the model using edit distance algorithm. The experimental results of phone recognition are shown in Table 3, which shows that the PER of the free phone recognition using the CNN-RNN-CTC is 27.75\%, the result means that the original model on these data does not perform well. After introducing a priori text information, the setence-to-phoneme and the setence-to-character reported on sed-mdd have performances of 16.06\% and 20.66\% respectively in our proposed model structure, we can find that taking phoneme recognition as the goal and using text phoneme sequence as input is helpful to improve the performance of phoneme recognition. Although we modified the prior phoneme sequence in data augmentation, the performance did not decrease, and randomly selected 10\% of the input phoneme sequence and randomly modified it according to the confusion-pairs(CP10\%) to achieve the best PER of 15.48\%.

\begin{table}[H]
\caption{Confusion matrices of most frequency misrecognized vowels and consonants}
\centering
\resizebox{\linewidth}{!}{%
\begin{tabular}{llllllll}
\hline
                   & \multicolumn{1}{c}{} & \multicolumn{6}{c}{Annotation}                                                               \\ \cline{3-8} 
                   &                      & aa            & ah            & ae            & eh           & ih            & iy            \\ \hline
CNN-RNN-CTC        & aa                   & 231           & 46            & 15            & 7            & 0             & 0             \\
                   & ah                   & 59            & 2035          & 37            & 51           & 72            & 16            \\
                   & ae                   & 20            & 32            & 630           & 62           & 5             & 0             \\
                   & eh                   & 1             & 45            & 75            & 544          & 43            & 3             \\
                   & ih                   & 0             & 105           & 29            & 49           & 1256          & 170           \\
                   & iy                   & 1             & 19            & 0             & 6            & 177           & 1097          \\
Attention-VC(10\%) & aa                   & \textbf{310}  & 12            & 14            & 1            & 0             & 0             \\
                   & ah                   & 33            & \textbf{2421} & 10            & 42           & 29            & 7             \\
                   & ae                   & 19            & 7             & \textbf{757}  & 19           & 0             & 1             \\
                   & eh                   & 1             & 15            & 32            & \textbf{694} & 14            & 1             \\
                   & ih                   & 0             & 13            & 8             & 19           & \textbf{1567} & 115           \\
                   & iy                   & 0             & 10            & 0             & 3            & 77            & \textbf{1214} \\ \hline
\multicolumn{8}{c}{most frequency misrecognized vowels}                                                                                  \\ \hline
                   &                      & d             & dh            & t             & sh           & s             & z             \\
CNN-RNN-CTC        & d                    & 1067          & 193           & 77            & 0            & 1             & 5             \\
                   & dh                   & 74            & 125           & 7             & 0            & 2             & 12            \\
                   & t                    & 173           & 14            & 1280          & 3            & 8             & 5             \\
                   & sh                   & 0             & 0             & 2             & 313          & 3             & 1             \\
                   & s                    & 8             & 4             & 12            & 10           & 1457          & 185           \\
                   & z                    & 3             & 3             & 4             & 0            & 97            & 228           \\
Attention-VC(10\%) & d                    & \textbf{1278} & 191           & 68            & 0            & 0             & 1             \\
                   & dh                   & 108           & \textbf{171}  & 6             & 0            & 2             & 17            \\
                   & t                    & 57            & 3             & \textbf{1433} & 1            & 5             & 2             \\
                   & sh                   & 0             & 0             & 0             & \textbf{322} & 7             & 0             \\
                   & s                    & 2             & 0             & 8             & 7            & \textbf{1466} & 138           \\
                   & z                    & 3             & 1             & 0             & 0            & 120           & \textbf{311}  \\ \hline
\multicolumn{8}{c}{most frequency misrecognized consonants}                                                                              \\ \hline
\end{tabular}%
}
\end{table}

\begin{table*}[]
\caption{Experiment result.}
\centering
\resizebox{\textwidth}{!}{%
\begin{tabular}{l|r|l|l|l|l|l|l|c|c|l|}
\cline{2-11}
 & \multicolumn{1}{c|}{\multirow{4}{*}{Models}} & \multicolumn{2}{c|}{canonicals}                                                                                                                                                               & \multicolumn{3}{c|}{mispronunciations}                                                                                                                                                                                                                       & \multirow{4}{*}{Recall} & \multirow{4}{*}{Precision} & \multirow{4}{*}{F-measure(\%)} & \multirow{4}{*}{PER(\%)} \\ \cline{3-7}
 & \multicolumn{1}{c|}{}                        & \multicolumn{1}{c|}{\multirow{3}{*}{\begin{tabular}[c]{@{}c@{}}True\\ Accept\end{tabular}}} & \multicolumn{1}{c|}{\multirow{3}{*}{\begin{tabular}[c]{@{}c@{}}False\\ Rejection\end{tabular}}} & \multicolumn{1}{c|}{\multirow{3}{*}{\begin{tabular}[c]{@{}c@{}}False \\ Accept\end{tabular}}} & \multicolumn{2}{c|}{True Rejection}                                                                                                                          &                         &                            &                     &                      \\ \cline{6-7}
 & \multicolumn{1}{c|}{}                        & \multicolumn{1}{c|}{}                                                                       & \multicolumn{1}{c|}{}                                                                           & \multicolumn{1}{c|}{}                                                                         & \multicolumn{1}{c|}{\begin{tabular}[c]{@{}c@{}}Correct \\  Diag.\end{tabular}} & \multicolumn{1}{c|}{\begin{tabular}[c]{@{}c@{}}Diag. \\ Error\end{tabular}} &                         &                            &                     &                      \\ \cline{2-11} 
 & \multicolumn{1}{c|}{CNN-RNN-CTC\cite{leung2019cnn}}             & 78.53\%(20194)                                                                              & 21.47\%(5520)                                                                                   & 25.22\%(1082)                                                                                 & 64.57\%(2072)                                                                  & 35.43\%(1137)                                                               & \textbf{74.78}\%                       & 36.76\%                          & 49.29                   & 27.75                \\ \cline{2-11} 
 & +Character-Attention                                   & 87.30\%(22449)                                                                              & 12.70\%(3265)                                                                                   & 37.68\%(1617)                                                                                 & 69.52\%(1859)                                                                  & 30.48\%(815)                                                                & 62.32\%                       & 45.02\%                          & 52.28                   & 20.66\                    \\ \cline{2-11}
 & +Phoneme-Attention                                   & \textbf{93.06\%(23929)}                                                                              & \textbf{6.94\%(1785)}                                                                                   & 49.59\%(2128)                                                                                 & 73.78\%(1595)                                                                  & 26.26\%(568)                                                                & 50.41\%                       & 54.79\%                          & 52.51                   & 16.06\                   \\ \cline{2-11} 
 & \multicolumn{10}{c|}{Performance of  different model architecture}                                                                                                                                                                                                                                                                                                                                                                                                                                                                                                                                              \\ \cline{2-11} 
 & +PS(10\%)                                    & 92.72\%(23841)                                                                              & 7.28\%(1873)                                                                                    & 45.07\%(1934)                                                                                 & 75.31\%(1775)                                                                  & 24.69\%(582)                                                                & 54.93\%                       & 55.72\%                          & 55.32               & 15.52                \\ \cline{2-11} 
 
 & +PS(15\%)                                    & 92.23\%(23715)                                                                              & 7.77\%(1999)                                                                                    & 43.18\%(1853)                                                                                 & 74.65\%(1820)                                                                  & 25.35\%(618)                                                                & 56.82\%                       & 54.95\%                          & 55.87               & 16.13                \\ \cline{2-11} 
 
 & +PS(20\%)                                    & 92.60\%(23810)                                                                              & 7.4\%(1904)                                                                                    & 44.63\%(1915)                                                                                 & 73.40\%(1744)                                                                  & 26.60\%(632)                                                                & 55.37\%                       & 55.51\%                          & 55.44               & 15.96                \\ \cline{2-11} 
 
 & +VC(10\%)                                    & 92.65\%(23825)                                                                              & 7.35\%(1889)                                                                                    & 43.88\%(1883)                                                                                 & 74.96\%(1805)                                                                  & 25.04\%(603)                                                                & 56.12\%                       & 56.04\%                          & \textbf{56.08}      & 15.58                \\ \cline{2-11} 
 & +VC(15\%)                                    & 92.27\%(23726)                                                                              & 7.73\%(1988)                                                                                   & 43.32\%(1859)                                                                                 & 74.63\%(1815)                                                                  & 25.37\%(617)                                                                & 56.68\%                       & 55.02\%                          & 55.84               & 15.71                \\ \cline{2-11} 
 & +VC(20\%)                                    & 91.77\%(23650)                                                                              & 8.03\%(2064)                                                                                   & 43.59\%(1870)                                                                                 & 73.90\%(1789)                                                                  & 30.42\%(812)                                                                & 56.42\%                       & 53.98\%                          & 55.17               & 16.33                \\ \cline{2-11} 
 & +CP(10\%)                                    & 92.83\%(23870)                                                                            & 7.17\%(1844)                                                                                   & 45.14\%(1937)                                                                                 & \textbf{75.45\%(1776)}                                                                  & \textbf{24.55\%(578)}                                                                & 54.86\%                       & \textbf{56.07}\%                          & 55.46               & \textbf{15.48}               \\ \cline{2-11} 
 & +CP(15\%)                                    & 92.62\%(23817)                                                                              & 7.38\%(1897)                                                                                   & 46.10\%(1978)                                                                                 & 73.07\%(1690)                                                                  & 26.93\%(623)                                                                & 53.90\%                       & 54.94\%                          & 54.42               & 15.91               \\ \cline{2-11} 
 
 & +CP(20\%)                                    & 91.95\%(23644)                                                                              & 8.05\%(2070)                                                                                   & \textbf{42.32\%(1816)}                                                                                 & 74.10\%(1834)                                                                  & 25.90\%(641)                                                                & 57.68\%                       & 54.46\%                          & 56.02               & 16.20               \\ \cline{2-11} 
 & \multicolumn{10}{c|}{Performance of different data augmentation with phoneme-attention model}                                                                                                                                                                                                                                                                                                                                                                                                                                                                                                                           \\ \cline{2-11} 
\end{tabular}%
}

\footnotesize
Note: The CNN-RNN-CTC is our baseline model and audio encoder. “Character-Attention” denotes that we use character sequences as sentence input on attention model our proposed. “Phoneme-Attention” denotes that we use phoneme sequences as sentence input. To modify the sentence input sequence, three data augmentation methods was used to here, "PS" means phoneme set based, "VC" means the vowels and consonants set base, "CP" means the confusing pairs based, we tried to modify separately 10\%,15\%,20\% of the phonemes in the input phoneme sequence

\end{table*}

The confusion matrices of most frequently misrecognized vowels and consonants by CNN-RNN-CTC and our best MD\&D model(Attention-VC10\%) are shown in Table 2.  Our model significantly obtains better results than CNN-RNN-CTC for all confusable consonants and vowels. In addition, we find that this tendency to confusable after the introducing the prior text still remains(i.e. /$d$/-/$t$/, /$z$/-/$s$/, etc.), and there is no obvious damage to the acoustic information.

\subsection{Results of MD\&D}

Following previous work \cite{leung2019cnn,li2016mispronunciation}, the hierarchical evaluation structure are used to measure the MD\&D system performance. The true accept(TA) and true rejection(TR) indicates predict correctly, while the false accept(FA) and false rejection(FR) indicates predict incorrectly. The true rejection(TR) is divided into correct diagnosis and diagnosis error. Three main evaluation indicators precision, recall and f-measure are calculated by the following formula:

\begin{equation}
    \mathrm{Precision = \frac{TR}{TR+FR}} \label{eq:precision}
\end{equation}
\begin{equation}
    \mathrm{Recall = \frac{TR}{TR+FA}} \label{eq:recall}
\end{equation}
\begin{equation}
    \mathrm{F-}\mathrm{measure = 2\frac{Precision*Recall}{Precision+Recall}} \label{eq:f1}
\end{equation}

For the model architecture, it can be found that our proposed phoneme-attention model improves the F-measure from 49.29\% to 52.51\% when comparing to CNN-RNN-CTC baseline, which is slightly better than character-attention model as well. We also notice a significant decrease in recall due to the large number of phonemes classified as input phonemes after the inclusion of prior text constraints, which is reflected in the TA reaching 93.06\%. In the actual application of the MD\&D, in order for learners to perceive the system as reliable, the TA needs to maintain a very high performance, which means learner's many correct pronunciation in the canonicals detected by the machine are correct.  


For the data augmentation, it can be seen that we achieved the best result in the experiment 56.08\% after using VC(10\%) method, which is an absolute improvement of 6.79\% compared with the baseline. In addition, all data augmentation experiments maintain high TA and f-measure performance. 

Through the analysis of the correct diagnosis and the incorrect diagnosis, we find that the main reason for poor recall performance are the decrease in the number of diagnosed error. What's more, we show a comparison of the results in detecting the type of mispronunciation as in Table 4, the error diagnosis of the three types is greatly reduced after adding a priori text information, especially the two types of substitution and deletion.

\begin{table}[H]
\centering
\caption{Detection result of mispronunciation types}
\resizebox{\linewidth}{!}{%
\begin{tabular}{|l|c|c|c|c|}
\hline
\multicolumn{2}{|l|}{}                                                                                                                        & \multicolumn{3}{c|}{Mispronunciations}                                                                                                                                                              \\ \hline
\multicolumn{2}{|c|}{Error types}                                                                                                             & Substitution                                                     & Insertion                                                       & Deletion                                                       \\ \hline
\multicolumn{2}{|c|}{Error nums}                                                                                                              & \begin{tabular}[c]{@{}c@{}}3154\\ (100\%)\end{tabular}           & \begin{tabular}[c]{@{}c@{}}219\\ (100\%)\end{tabular}           & \begin{tabular}[c]{@{}c@{}}918\\ (100\%)\end{tabular}          \\ \hline
\multirow{2}{*}{\begin{tabular}[c]{@{}l@{}}Correct\\ Diag.\end{tabular}} & CNN-RNN-CTC                                                        & \begin{tabular}[c]{@{}c@{}}1475\\ (46.77\%)\end{tabular}         & \begin{tabular}[c]{@{}c@{}}98\\ (44.75\%)\end{tabular}          & \begin{tabular}[c]{@{}c@{}}499\\ (54.36\%)\end{tabular}        \\ \cline{2-5} 
                                                                         & \begin{tabular}[c]{@{}c@{}}Phoneme-Attention\\ (VC10\%)\end{tabular} & \begin{tabular}[c]{@{}c@{}}1233\\ (39.09\%)\end{tabular}         & \begin{tabular}[c]{@{}c@{}}90\\ (41.10\%)\end{tabular}          & \begin{tabular}[c]{@{}c@{}}482\\ (52.50\%)\end{tabular}        \\ \hline
\multirow{2}{*}{\begin{tabular}[c]{@{}l@{}}Diag.\\ Error\end{tabular}}   & CNN-RNN-CTC                                                        & \begin{tabular}[c]{@{}c@{}}829\\ (26.28\%)\end{tabular}          & \begin{tabular}[c]{@{}c@{}}71\\ (32.42\%)\end{tabular}          & \begin{tabular}[c]{@{}c@{}}237\\ (25.82\%)\end{tabular}        \\ \cline{2-5} 
                                                                         & \begin{tabular}[c]{@{}c@{}}Phoneme-Attention\\ (VC10\%)\end{tabular} & \textbf{\begin{tabular}[c]{@{}c@{}}493\\ (15.63\%)\end{tabular}} & \textbf{\begin{tabular}[c]{@{}c@{}}47\\ (21.46\%)\end{tabular}} & \textbf{\begin{tabular}[c]{@{}c@{}}63\\ (6.86\%)\end{tabular}} \\ \hline
\end{tabular}%
}

\end{table}

\section{Conclusions}
In this paper, we present a fully text-dependent end to end model for mispronunciation detection and diagnosis (MD\&D). The model does not need to use any forced alignment information and only requires audio and prior phoneme sequences to be available for training, which makes the model easier to train and a wider range of applications. Furthermore, three easy data augmentation techniques are used to make up for the shortcomings of our proposed model. By doing experiments on public corpora, we demonstrate that our proposed method shows more effective performance than the baseline system on the TA and F-measure. 
In the future, considering that we don’t have a lot of training data, the pre-trained model and unsupervised learning will be investigated to improve the performance with our model.

\clearpage
\bibliographystyle{IEEEtran}
\bibliography{references.bib}


\end{document}